\documentclass[a4paper]{article}

\usepackage[final]{nips_2018}

\usepackage[english]{babel}

\usepackage{subfig}
\usepackage[toc,page]{appendix}
\usepackage{color}
\usepackage{stmaryrd}
\usepackage{amsmath}
\usepackage[utf8]{inputenc} 
\usepackage[T1]{fontenc}    
\usepackage{hyperref}       
\usepackage{url}            
\usepackage{booktabs}       
\usepackage{amsfonts}       
\usepackage{nicefrac}       
\usepackage{microtype}      
\usepackage{amsmath}
\usepackage{amssymb}
\usepackage{graphicx}
\usepackage{stmaryrd}
\usepackage{stackengine}
\usepackage[]{algorithm2e}

\definecolor{orange}{rgb}{1,0.5,0}

\newcommand{\ie}{\emph{i.e. }}

\newcommand{\X}{\mathcal{X}}

\newcommand{\Y}{\mathcal{Y}}

\DeclareMathOperator*{\argmax}{arg\,max}
\DeclareMathOperator*{\argmin}{arg\,min}

\title{Computer-Assisted Fraud Detection,\\ From Active Learning to Reward Maximization}
\author{
  Christelle Marfaing \\
  Lydia Solutions\\
  \texttt{christelle.marfaing@lydia-app.com} \\
  \And
  Alexandre Garcia\\
  LTCI T\'el\'ecom ParisTech\\
  Universit\'e Paris Saclay\\
  \texttt{garcia@telecom-paristech.fr} \\
}

\begin{document}
\maketitle

\begin{abstract}
The automatic detection of frauds in banking transactions has been recently studied as a way to help the analysts finding fraudulent operations. Due to the availability of a human feedback, this task has been studied in the framework of active learning: the fraud predictor is allowed to sequentially call on an oracle. This human intervention is used to label new examples and improve the classification accuracy of the latter. Such a setting is not adapted in the case of fraud detection with financial data in European countries. Actually, as a human verification is mandatory to consider a fraud as really detected, it is not necessary to focus on improving the classifier. We introduce the setting of 'Computer-assisted fraud detection' where the goal is to minimize the number of non fraudulent operations submitted to an oracle. The existing methods are applied to this task and we show that a simple meta-algorithm provides competitive results in this scenario on benchmark datasets.
\end{abstract}
\section{Introduction}
The task of automatic fraud detection has been mainly studied under the framework of imbalanced binary classification \citep{bhattacharyya2011data}. Given the description of a transaction $x$, the goal is to predict a binary label $y \in \{1,0\}$ indicating whether this transaction is fraudulent or not. The main difficulties arising in fraud detection highlighted earlier in \citep{bolton2002statistical} include among others
\begin{itemize}
    \item The strong imbalance between the output labels. Indeed fraudulent behaviors are assumed to be rare and thus harder to find. Previous work have proposed different solutions that help building efficient predictors in the case of imbalanced classification. Such approaches mainly consist in introducing instance reweighting or bootstrap based schemes \citep{chawla2004special} in order to transform the imbalanced learning problem in a related balanced problem on which learning can be done with on the shelves predictors.
    \item The large amount of unlabeled data in regards to labelled data advocates the use of methods that can scale on large datasets and that generalize well.
\end{itemize}
In the case of fraud detection in financial transactions, these properties have been highlighted in work involving both supervised \citep{owen2007infinitely} and unsupervised \citep{damez2012dynamic} learning approaches where the problem of handling large datasets is specifically studied. Whereas in this setting the user assumes that he has enough labeled samples to confidently follow the decision of a learned predictor, another line of work relies on an active learning procedure that consists in optimizing the accuracy of a predictor by iteratively labelling a set of well chosen samples \citep{Carcillo2018StreamingAL,zhang2018online}. The main objective of this approach is to minimize the amount of work necessary to build a correctly performing classifier since obtaining reliable labels is an expensive operation. A common feature of the existing active learning strategies is the selection of examples that keep a balanced pool of labeled samples wherever it is in label or in space \citep{ertekin2007learning}. Indeed, training a predictor with imbalanced data is known to affect its performance while incorporating some scalability issues due to the difficulty to handle in memory the labeled examples of the majority class. This online re-balancing process solves thus the two issues raised above at the same time.  

In the standard active learning framework, the true labels are sequentially queried to an oracle and a good active learning strategy should be able to provide good classification performance on some new sample while doing as few oracle queries as possible. Whereas previous work in the context of fraud detection have focused on optimizing some metrics computed on a test set based on the resulting classifier predictions, we argue that in many practical applications with financial data, this setting is not adapted since it does not rely on the right metric. Actually, Due to the article 22 of the GPDR european regulation - automated individual decision-making, including profiling -, engaging some legal pursuits and sanctions against a fraudulent user requires a human verification of the corresponding decision \citep{eu-269-2014}. Since the fitted classifier will never be used without requesting an oracle, it is not desirable for the fitted classifier to outperform on a held-out dataset. It is preferable, in this configuration, to minimize the number of verifications that corresponds to non fraudulent operations and thus maximize the number of discovered and treated frauds over time. This setting differs from the active learning setting in the fact that the goal is not to build the best classifier over a given horizon but instead to recommend as many fraudulent objects as possible to the oracle. In the next sections we present each framework and stress on their similarities and differences as well as the consequences in terms of adapted strategy. 
\section{Mathematical Settings}
\subsection{The Active Learning Setting}
In this section we assume that 
we have access to a sample $\mathcal{D} = \{x,y \sim \mathcal{P}_{\X \times \Y} \}$ where $x$ are input feature representations and $y\in \{0,1\}$ binary labels indicating whether the transaction $x$ is a fraud ($y=1$) or not. This sample is partitioned into an active set $\mathcal{D}$ and a finite testing set $\mathcal{D}'$. The active set is again partitioned in a labeled set $\mathcal{D}_l$ and an unlabeled set $\mathcal{D}_u$ that evolve over time since querying the label of an example make it move from $\mathcal{D}_u^t$ to $\mathcal{D}_l^t$ at each iteration $t$. Initially, the labels in the labeled set are only available for a fraction of the data $\mathcal{D}_{l}^{1} = \{ x_i, y_i\}_{i \in 1,\ldots, n} $. We suppose additionally that we have access to an active learning strategy $H_t(\mathcal{D}_l^t)$ \ie a function based on the current labeled sample which returns the next unlabeled point that will be provided to the oracle. Example of such strategies are detailed in section \ref{strategies}. The active learning procedure can then be divided in the following steps:
\begin{enumerate}
    \item Based on the current labeled dataset $\mathcal{D}_l^{t}$, build a predictor $g_t$
    \item Choose an unlabelled point $x$ based on $H_t(\mathcal{D}_l^{t},g_t)$ for which we want to obtain the label.
    \item Query the corresponding label $y$ to an oracle
    \item Update $\mathcal{D}_l^{t+1} = \mathcal{D}_l^{t} \bigcup \; (x,y)$, and $\mathcal{D}_u^{t+1} = \mathcal{D}_u^{t} \setminus \; (x,y)$ 
    \item Increment $t$ and repeat from (1) until $t=T$.
\end{enumerate}
The performance of an active learning strategy can be measured thanks to the performance of $g_t$ on
the testing set. For a non-negative performance measure $m:\Y \times \Y \rightarrow \mathbb{R}^+$, the goal is to find the strategy that maximizes $\sum_{(x,y)\in \mathcal{D}'} m(g_t(x),y)$ for all the time steps $t\in \{1,\ldots,T\}$.
\subsection{Active learning strategies}
\label{strategies}
A large body of work has focused on designing active learning strategies that take into account some properties of the data or some specificities of the underlying class of predictors to optimize. Thus \citep{ertekin2007learning} focuses on learning on the border using SVM properties, and \citep{zhang2016online} uses the distance notion introduced by the SVM hyperplane to define a way to query points to label. On the other hand, strategies can be defined without relying on some properties of the underlying predictor but only take advantage of their ability to produce class wise probability estimations. Such strategies 
can be grouped into two categories :
\begin{itemize}
\item 
Unitary methods (base methods): Uncertainty Sampling, Random sampling. This type of method rely on a single hypothesis explaining the insufficient performance of the predictor. Based on this hypothesis, a sampling method is proposed. In the case of Uncertainty sampling the hypothesis is that the most important are where the probabilities estimated by the model itself have a high variance \cite{lewis1994heterogeneous,cohn1995active}. In practice the strategy will tend to select samples in zones that are at the known frontier of two distinct classes. In the case of Random sampling, the strategy ignores the learned predictor and makes no hypothesis on the evolution of its performance with respect to the chosen labeled points.
\item Adaptive methods: While unitary methods have been designed with the idea of choosing samples that optimize a single criterion,   \citep{hsu2015active} proposes a meta-algorithm that chooses the best unitary method to use at each time step in order to maximize a specifically designed reward function (Weighted accuracy computed on the points submitted to the oracle). Note for example that different uncertainty sampling approaches could be built based upon different probability estimations of the output labels and the adaptive approach would choose at each time step which unitary strategy should be chosen. Similarly,  \citep{Konyushkova2017LearningAL} fits a model able to predict the expected increase of a test metric. Then, the point picked by the algorithm is the one that has the greatest expected reward in the so-called metrics.
\end{itemize}
Now we turn to the presentation of our framework that differs from the active learning one.
\subsection{The Reward Maximization framework}
\label{framework}
We now propose a new setting that intends to simulate more appropriately the real-life constraints. The goal is no longer to optimize a metric evaluated on a holdout dataset but instead to iteratively retrieve only the examples corresponding to the class 1 (fraudulent operation) to the oracle. The available data is thus only partitioned into a labeled set $\mathcal{D}_t$ and unlabeled set $\mathcal{D}_t'$ such that $\mathcal{D} = \mathcal{D}_t \bigcup \mathcal{D}_t'$. Suppose that we can build a strategy $H_t$ that returns an unlabelled example. Given a non-negative reward function $r:\Y\times\Y \rightarrow \mathbb{R}^+$ the goal is then to find the strategy that maximises the cumulated reward:
\begin{equation}
\sum_{(x,y)\in \mathcal{D}_t} r(g_t(x),y)
\end{equation}
The reward can take into account the amount of money contained in a fraudulent transaction. When this information is not available, we can simply provide a unitary reward when a fraud is identified :
\begin{equation}
r(y,f(x)) = \begin{cases} 1 \; \text{if} \; y = f(x) =1 \\
0 \text{ else}
\end{cases}
\end{equation}
At each time step, the optimal strategy $H_t$ would return an element of $\mathcal{D}_t'$ in the set of highest expected reward.
$
\label{opt_sol}
    x^\star \in \argmax_x p(y = 1 | x)
$,
where $p(y|x)$ is the true conditional distribution of the data. 
Since the conditional probability is not directly available, it is instead estimated by a function $\hat{p}_t$ taken in a hypothesis class $\mathcal{C}$ and learned on the labeled sample $\mathcal{D}_t$:
\begin{equation}
    \hat{p}_t = \argmin_{\hat{p} \in \mathcal{C}} \sum_{x,y \in \mathcal{D}_t} l(\hat{p}(x),y) + \Omega(\hat{p})
\end{equation}
Where $l$ is a loss function penalizing wrong predictions of $y$ and $\Omega$ a penalty function enforcing the choice of regular candidates. 

In the case where we want to compute class probabilities, one can choose the cross entropy loss function:
\begin{equation}
    \hat{p}_t = \argmin_{\hat{p}\in \mathcal{C}} \sum_{x,y \in \mathcal{D}_t} - y \log(\hat{p}(x))
\end{equation}
This type of probability estimators are well known and can be parameterized by a linear (logistic regression) or a non linear model (neural networks). Different choices of loss and parameterization lead to different class of predictors that may be used to construct $\mathcal{C}$ (Gaussian Processes, Random Forests, Boosting based algorithm). Up to this point we have provided an approximation of $p(y|x)$ based on the $\mathcal{D}_t$ sample only. This has two consequences:
\begin{itemize}
\item Based on the knowledge of $\hat{p}$, the $x$ values proposed to the oracle will be the one with the highest probability of finding the label $1$. For a correctly regularized predictor, these points will be the one located close to already detected frauds. By analogy with the bandit litterature \cite{audibert2009exploration}, this step can be seen as an exploitation phase where the strategy relies on its estimation of the expected rewards to pick the arm that will give a gain with the highest probability among all the possible candidates.
\item When there are unlabeled parts of the space $X$ containing some objects labeled $1$ or when the ones we already found have been exhausted, then a good strategy needs to quickly explore the space to find new instances labeled $1$. During this step, instead of choosing the $x$ that maximizes the corresponding reward, we try to find the one that gives the most information to $\hat{p}$. Once again it is analogous to the exploration phase in the bandit literature. 
\end{itemize}
\section{A Computer-Assisted Fraud Detection Algorithm (\texttt{CAFDA})}
\label{sec:computeraided}
The two steps of exploration / exploitation presented previously can be mixed in a simple algorithm that in practice works surprisingly well on benchmark datasets for the task of computer-assisted fraud detection. It is inspired by the EXP4.P bandit algorithm \cite{beygelzimer2011contextual} which maintains a set of probability of picking each of the possible strategy and update them according to the reward received. 

Similarly to \cite{beygelzimer2011contextual,hsu2015active}, we suppose that we have access to a set of $K$ active learning algorithm that provide an advice vector $\mathbf{\xi}$ of the size of the unlabelled set which contains the probability of querying each example. We additionally maintain a vector $w \in [0,1]^K$ that indicates the probability of using each strategy and choose two update parameters $K_0$, $K_1$ which control the variation of $w$ depending on the rewards received. We also introduce $P_{min}$ and $P_{max}$ two threshold levels on the probabilities stored in $w$ that are used to reduce the time necessary to switch quickly from one best current strategy (of index $i$ with high $w_i$ value) to another as the number of iterations increases.
In order to maximize our custom reward, we propose the following fraud detection algorithm (\texttt{CAFDA}):
 \begin{algorithm}[H]
 \KwData{Labeled set $\mathcal{D}_l^{1}$ and unlabeled set $\mathcal{D}_u^{1}$
 }
 \KwResult{Sequence of rewards $(r_t)_{t\in\{1,\ldots,T\} }$, final labelled set $\mathcal{D}_l^T$}
 \textbf{Initialization:} Set initial probability of sampling each strategy $w_i=\frac{1}{K}$\;
 \For{$t$ in $\{1,\ldots,T\}$}{
  \textbf{Pick a strategy} $i \in\{1,\ldots,K\}$ according to the distribution $w$\;
  \textbf{Sample} the next point $x_j$ for which we want a query according to $\mathbf{\xi_i}$\;
  \textbf{Query} the label $y_j$ to the oracle\;
  \textbf{Receive a reward} $r_t$ according to $y_j$\;
  \textbf{Update the sets:} $\mathcal{D}_l^{t+1}= \mathcal{D}_l^{t}\bigcup (x_j,y_j)$ and $\mathcal{D}_u^{t+1}= \mathcal{D}_u^{t}\setminus (x_j,y_j)$\;
  \textbf{Update the probabilities} $w$ according to the following heuristic: \;
  \eIf{r = 1}{
  $w_i = \min(K_1 w_i,P_{max}) $ \;
  }{
  $w_i = \max(K_0 w_i,P_{min})$ \;
  }
  $\forall j \neq i, \; w[j]=\max(\min(w[j],P_{max}),P_{min})$   \;
  $w = \frac{w}{\sum_{i=1}^K w_i}$ \;
  \textbf{Update the strategies} using $\mathcal{D}_l^{t+1}$ \;
 }
 \caption{Heuristic based procedure for computer aided fraud detection (\texttt{CAFDA}}
\end{algorithm}
The main difference with \cite{hsu2015active} is the use of the $w$ update heuristic. In the original paper, the reward update scheme is chosen to optimize the accuracy of the resulting predictor on a held-out dataset which differs from our reward based only on the $1$ label found. Concerning the update, $EXP4.P$ has been designed to achieve optimal regrets in a stationary context which is not the case here. By choosing carefully $K_0$, $K_1$, $P_{min}$, $P_{max}$, \texttt{CAFDA} retrieves competitive results that we detail in section (\ref{sec:Experiments}).
%
%
\section{Experiments}
\label{sec:Experiments}
We simulate the framework described in section \ref{framework} in the following way.
Given an imbalanced fraud dataset containing $p$ frauds, we first sample a small fraction of the points that will constitute an initial labeled set and then iteratively select an unlabelled point which is shown to the oracle. If this point is labeled $1$, a reward of $1$ is gained and we display the cumulated reward over the time. We compare \texttt{CAFDA} against some baselines and state of the art active learning strategies:
\begin{itemize}
\item \texttt{base}: Use the predictor trained only once on the initial labeled set and perform the exploitation phase only at each time step: $x^\star \in \argmax_x \hat{p}_t(y = 1 | x)$ 
\item \texttt{base\_refit}: Same as base but the predictor is retrained on $\mathcal{D}_l^t$ at each timestep.
\item \texttt{random}: The point queried is picked randomly in the unlabeled set $\mathcal{D}_u^t$
\item \texttt{us} (uncertainty sampling): The point queried is the one of maximal uncertainty for the predictor \ie $\min_x(|\mathbb{P}(y=1|x) - \mathbb{P}(y=0|x)|)$
\item \texttt{lal\_independent} (Learning Active learning with an independent strategy): The point queried is the one of the maximal expected improve in a choosen loss. The expected improve is the prediction of a model fitted on a synthetic dataset. In the independent strategy, a Monte Carlo procedure is simulated to query randomly some points and associate them with an improve in the loss. \citep{Konyushkova2017LearningAL} 
\item \texttt{lal\_iterative}(Learning Active learning with an iterative strategy): This algorithm differs from the previous one only by the way the synthetic dataset is constructed. Actually, the points are queried in order to minimize the selection bias.
\item \texttt{albl} (Active Learning By Learning): A multi-armed bandit chooses among multiple active learning strategies at each time step in order to maximise an expected cumulated reward which is a weighted accuracy on the already queried point $\mathcal{D}_l^t$.
\end{itemize}

As base strategies for \texttt{CAFDA}, we take 5 strategies (base, base\_refit, random, lal\_independent, lal\_iterative) and exclude ALBL as it is also a meta-algorithm. For all the scenarios, $K_0=0.8$, $K_1=1.2$, $P_{min}=0.001$ and $P_{max}=0.95$. 

The different methods are compared in two scenarios:
\begin{enumerate}
\item The active learning is run during the entire experiment. In this experiment, we empirically show that active learning methods do not maximize the cumulated reward we defined.
\item The active learning algorithm is run for 100 steps, then the resulting classifier is used to select the points labeled $1$ with the highest probability. Here we aim at showing that early exploration using an active learning strategy is not even helping in the long run.
\end{enumerate}
For all our experiments, we used a Random Forest classifier as the base probability estimator and selected the hyperparameters by cross-validation on the initially labeled training set.

We display results obtained with 3 standard benchmark anomaly detection datasets since they share the imbalance property of financial fraud detection databases and are freely available. 
\begin{table}[ht]
\centering
\caption{Properties of the datasets}
\begin{tabular}{llll}
\hline
                   & shuttle & covtype & credit card \\ \hline
Number of samples  &  85849       &    295541     &    284807         \\ 
Input dimension    &       10  &      55   &         31    \\ 
Anomaly proportion &    7.2\%     &      4.1\%   &       0.17\%      \\ \hline
\end{tabular}
\end{table}
In all experiments, we first sample an initial labeled dataset. This initial set is 1\% for all the datasets. In the case of the covtype dataset, instead of using the full dataset, we worked with a sample of $10 000$ examples in order to keep a fairly low number of 'fraud' initially observed.
\subsection{Scenario 1}
\begin{figure}[t]
    \centering
    \subfloat[creditcard scenario 1]{{\includegraphics[width=8cm]{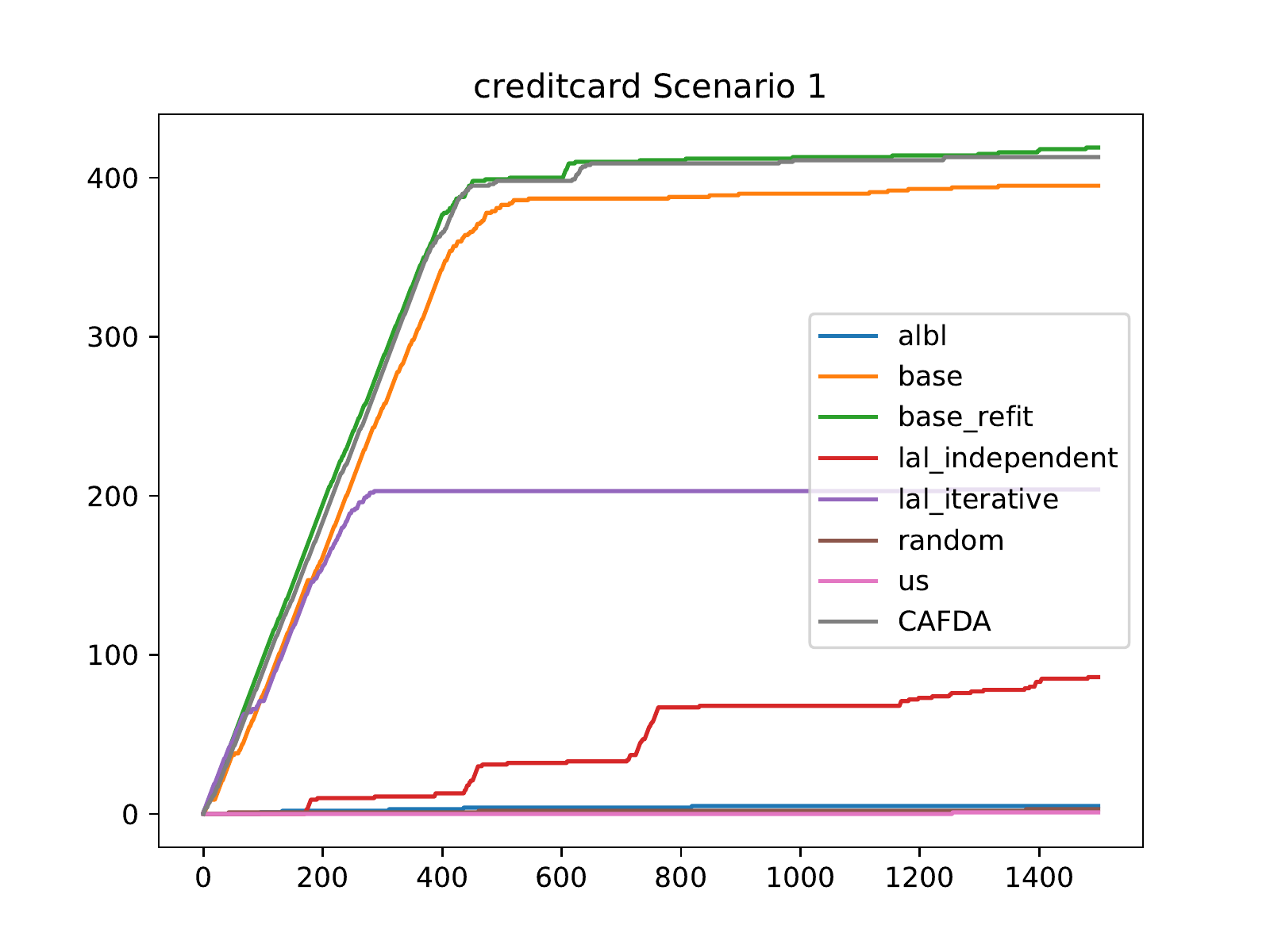} }}
    \subfloat[covtype scenario 1]{{\includegraphics[width=8cm]{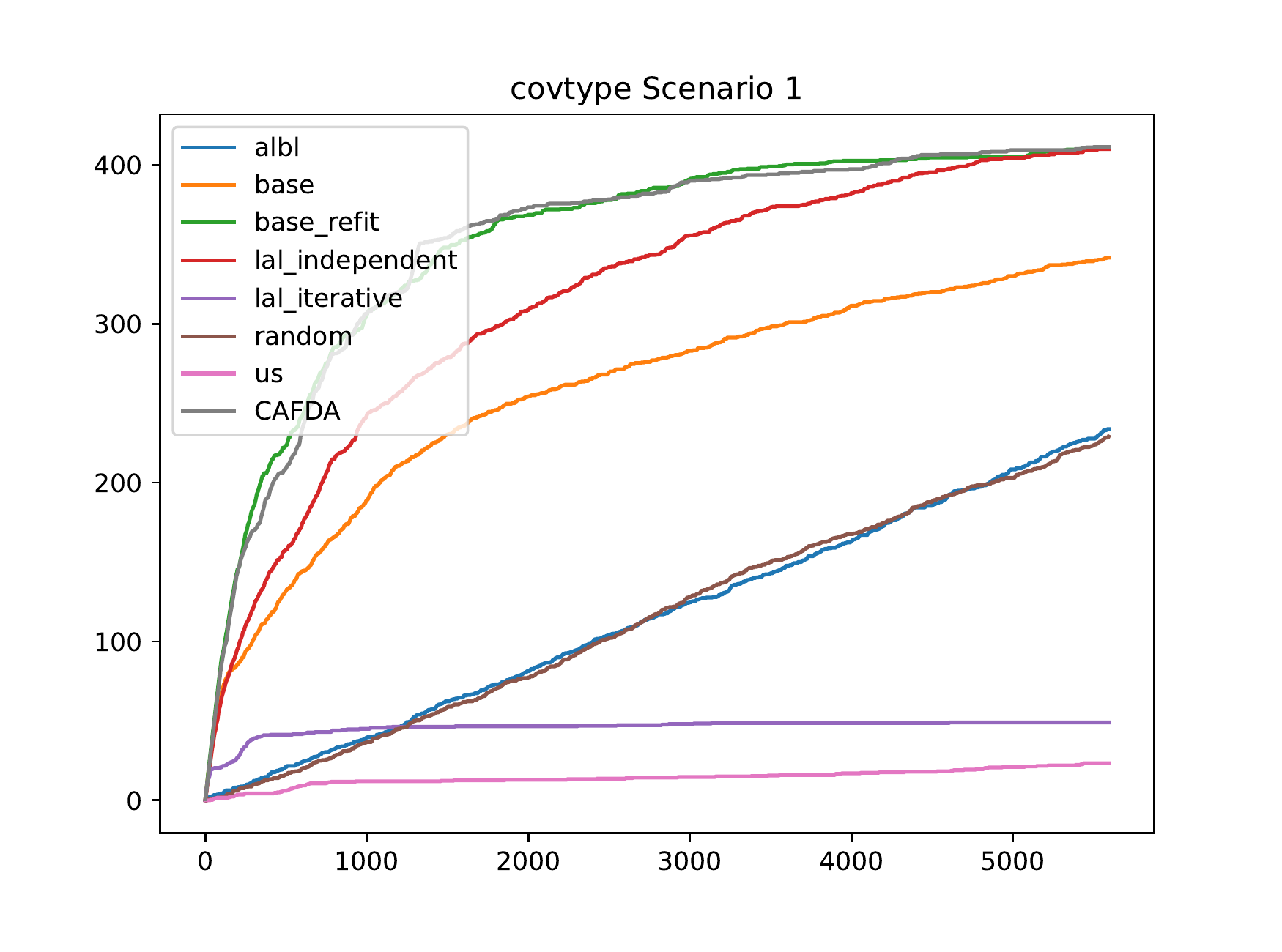} }}
    \qquad
    \subfloat[shuttle scenario 1]{{\includegraphics[width=8cm]{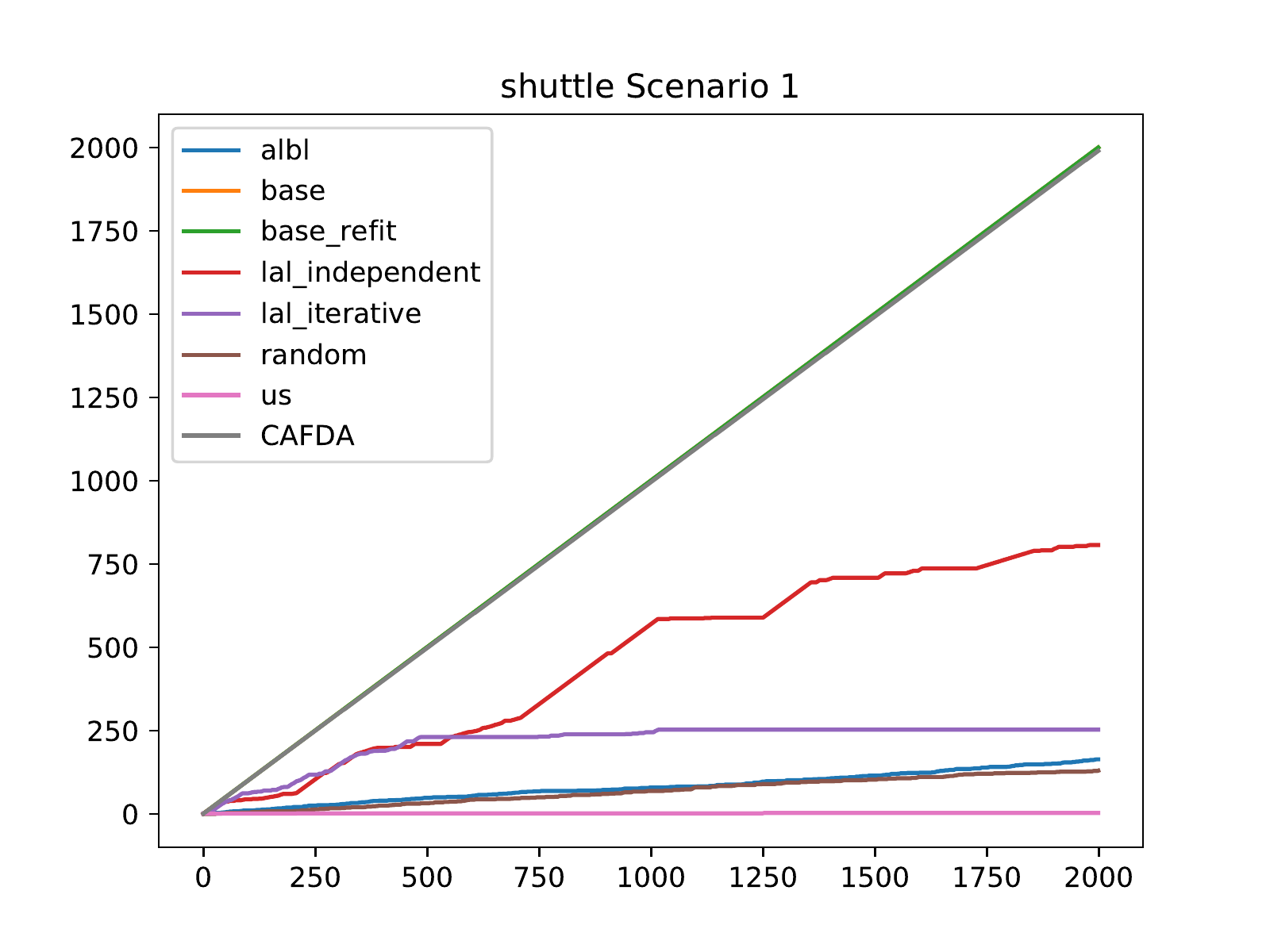} }}
    \caption{Scenario 1 cumulated rewards}%
    \label{fig_1:scenario1}%
\end{figure}

The figure \ref{fig_1:scenario1} shows the cumulated reward over time obtained with each strategy.

The best results are obtained using \texttt{CAFDA}, \texttt{base} and \texttt{base\_refit} with a slight improvement given to the model that retrain their underlying model on the creditcard dataset. As expected, the active learning strategies do not specifically try to provide $1$ labels to the oracle which explains their behavior. Now we test experimentally whether an early active learning based exploration can provide a benefit in a subsequent exploitation phase.
\subsection{Scenario 2}
\begin{figure}[]
    \centering
    \subfloat[covtype scenario 2]{{\includegraphics[width=8cm]{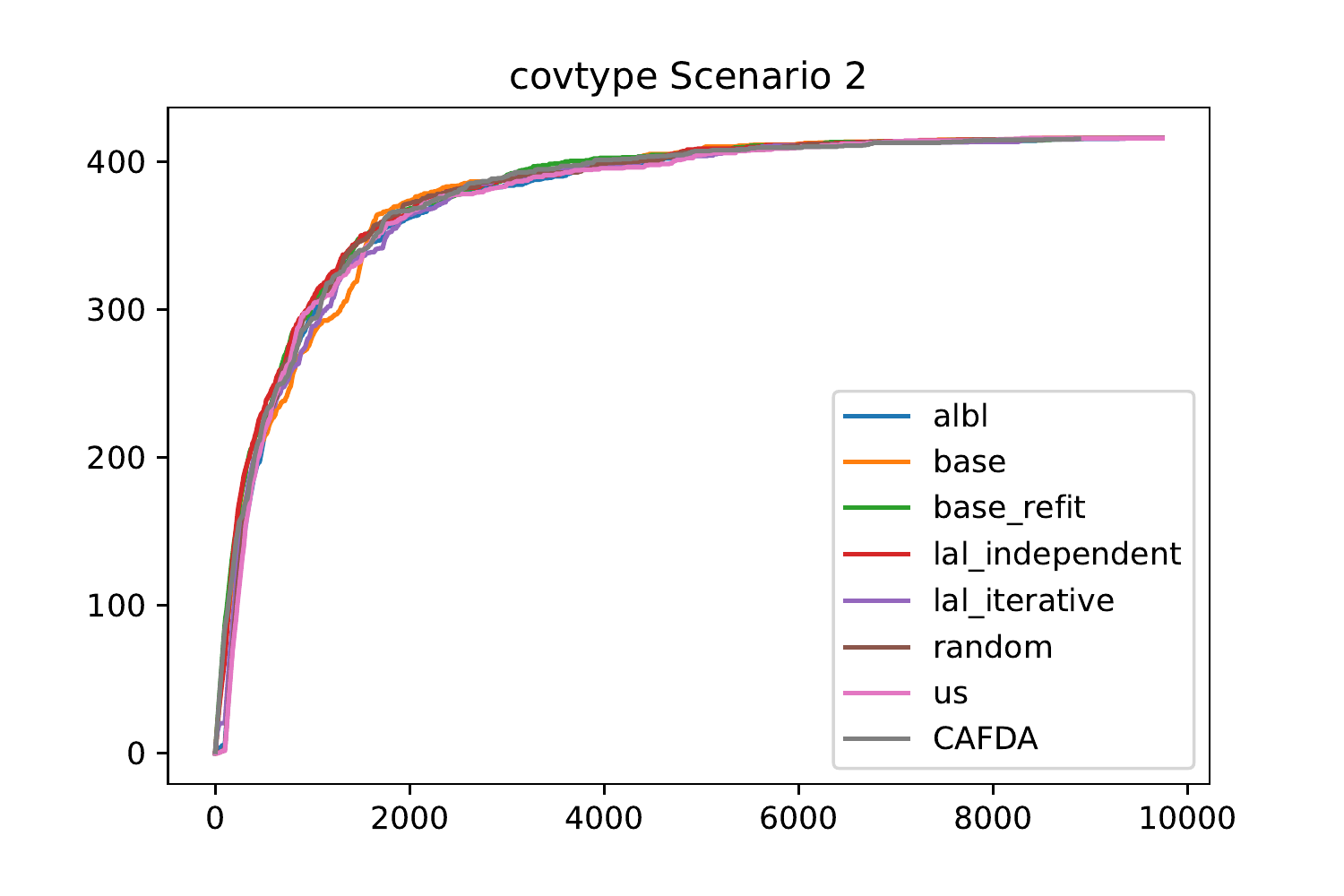} }}
    \subfloat[covtype scenario 2 first iterations]{{\includegraphics[width=8cm]{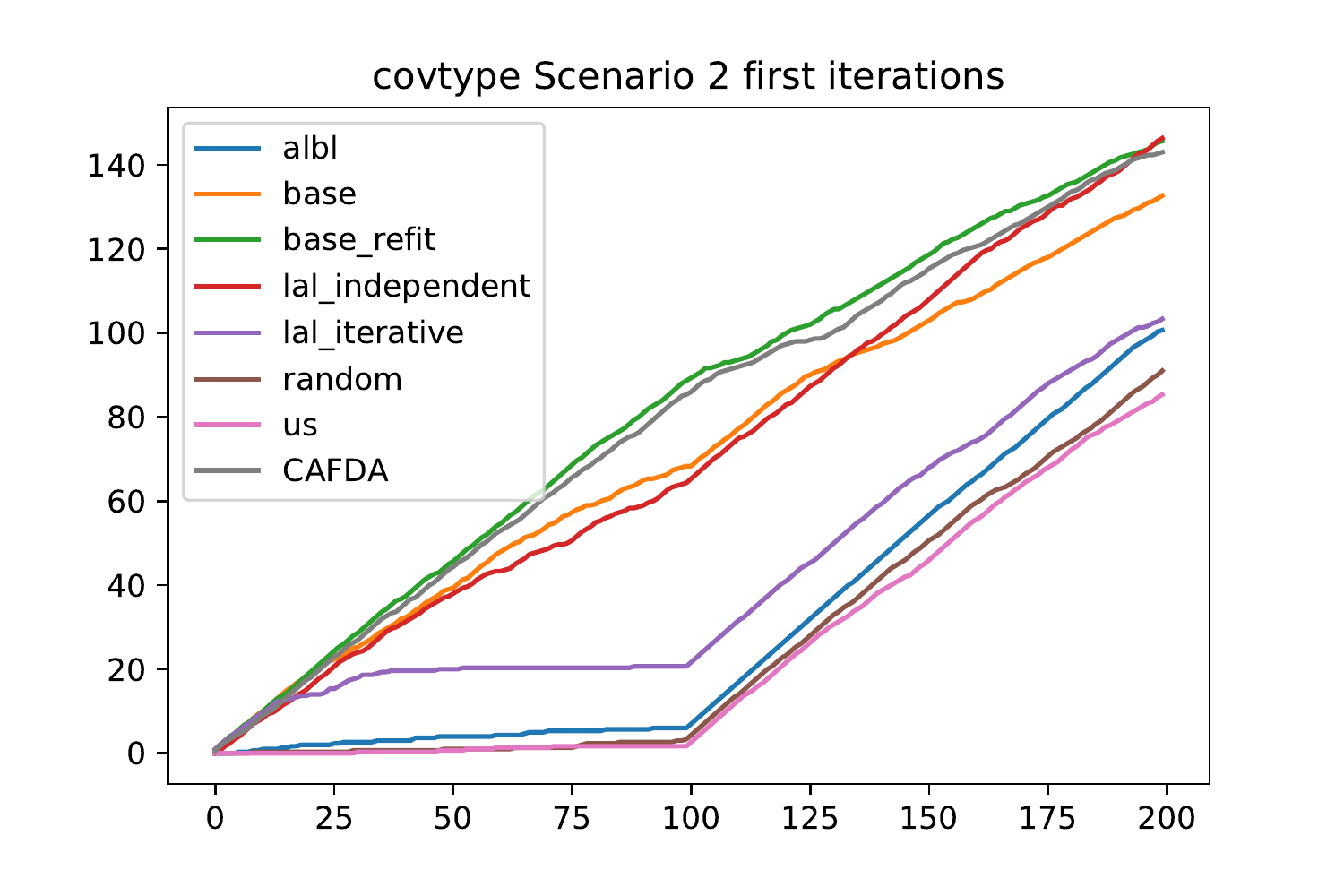} }}
    \qquad
    \subfloat[shuttle scenario 2]{{\includegraphics[width=8cm]{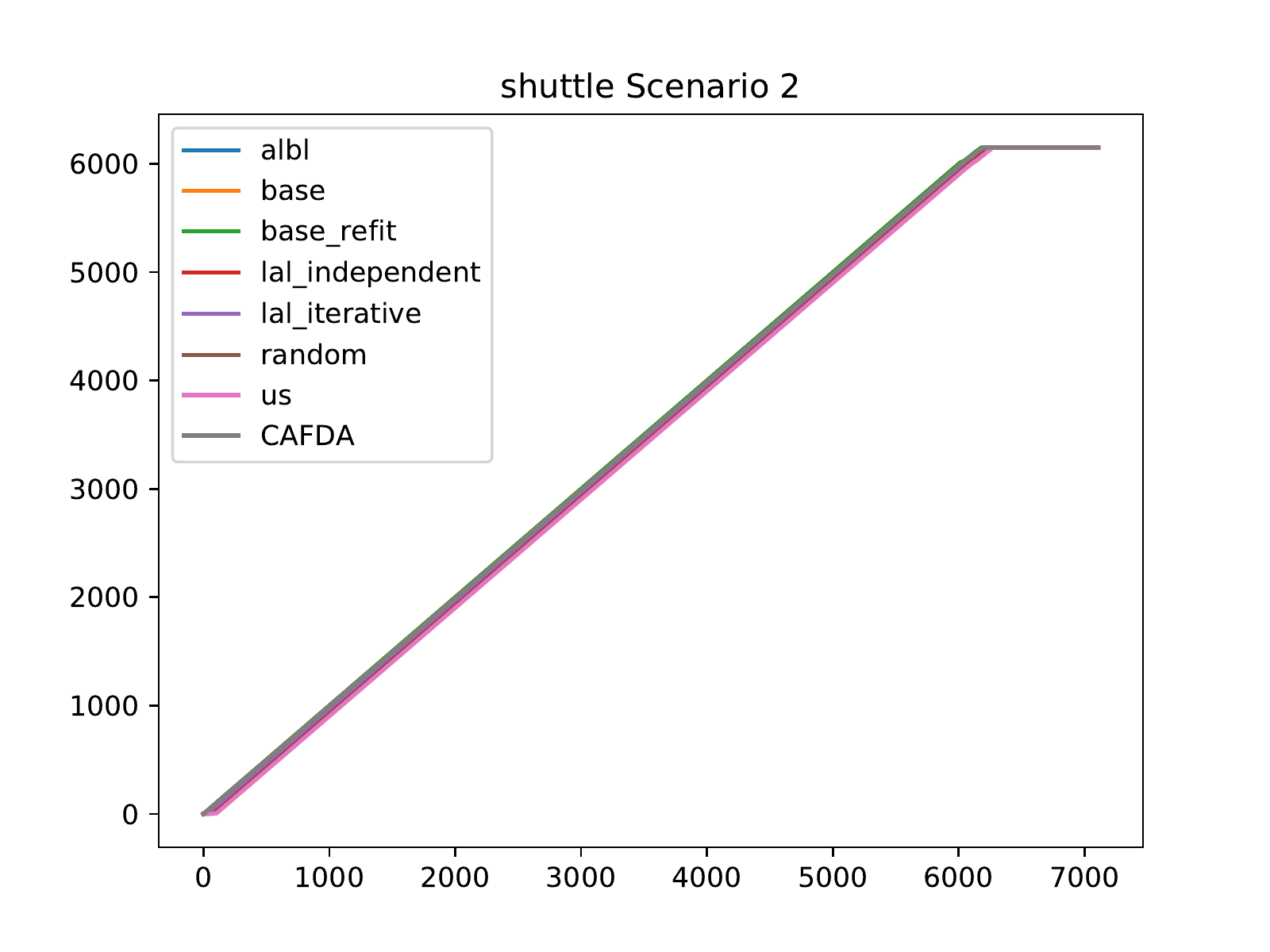} }}
    \subfloat[shuttle scenario 2 first iterations]{{\includegraphics[width=8cm]{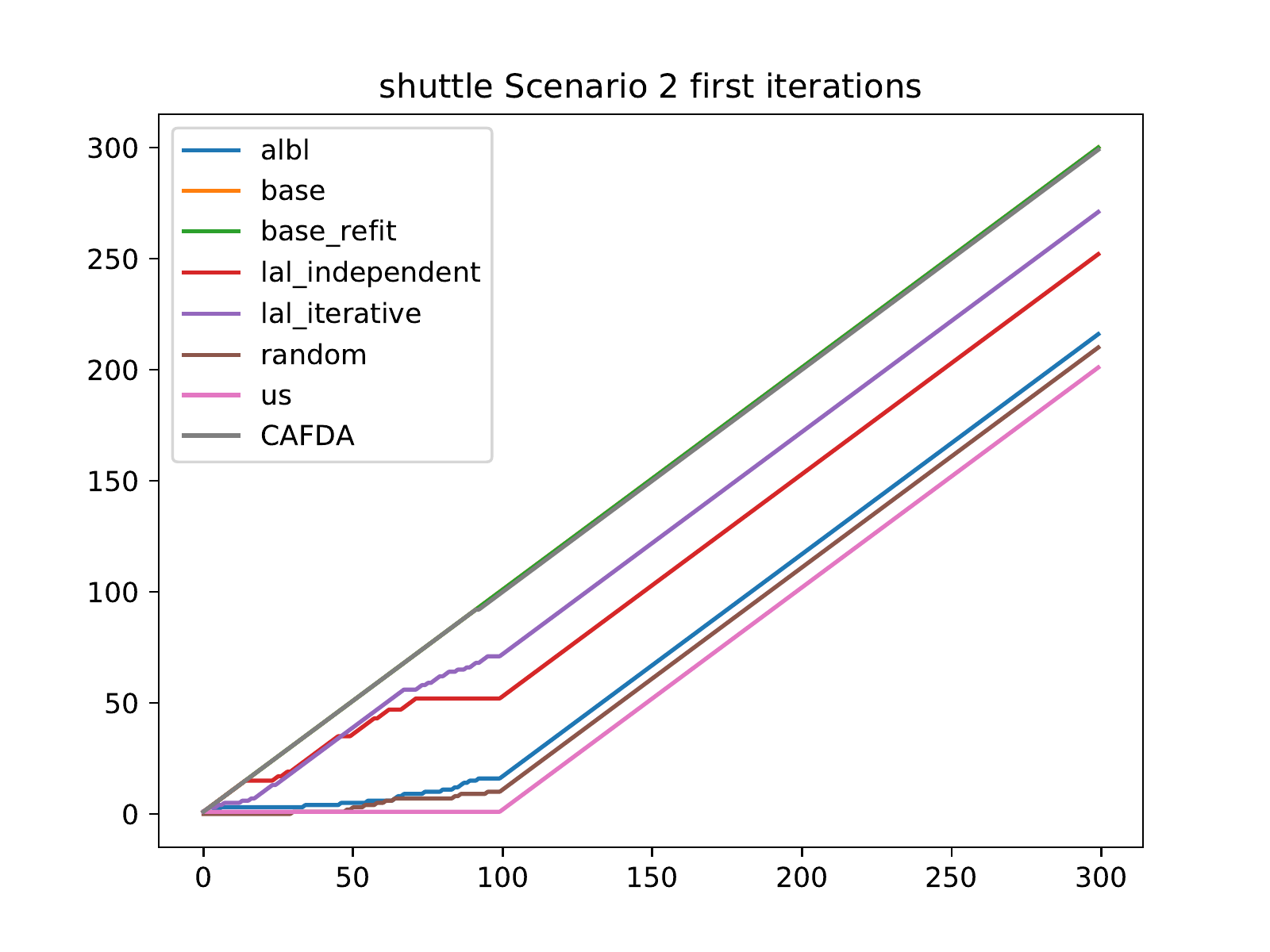} }}
    \qquad
    
    \subfloat[credit card scenario 2]{{\includegraphics[width=8cm]{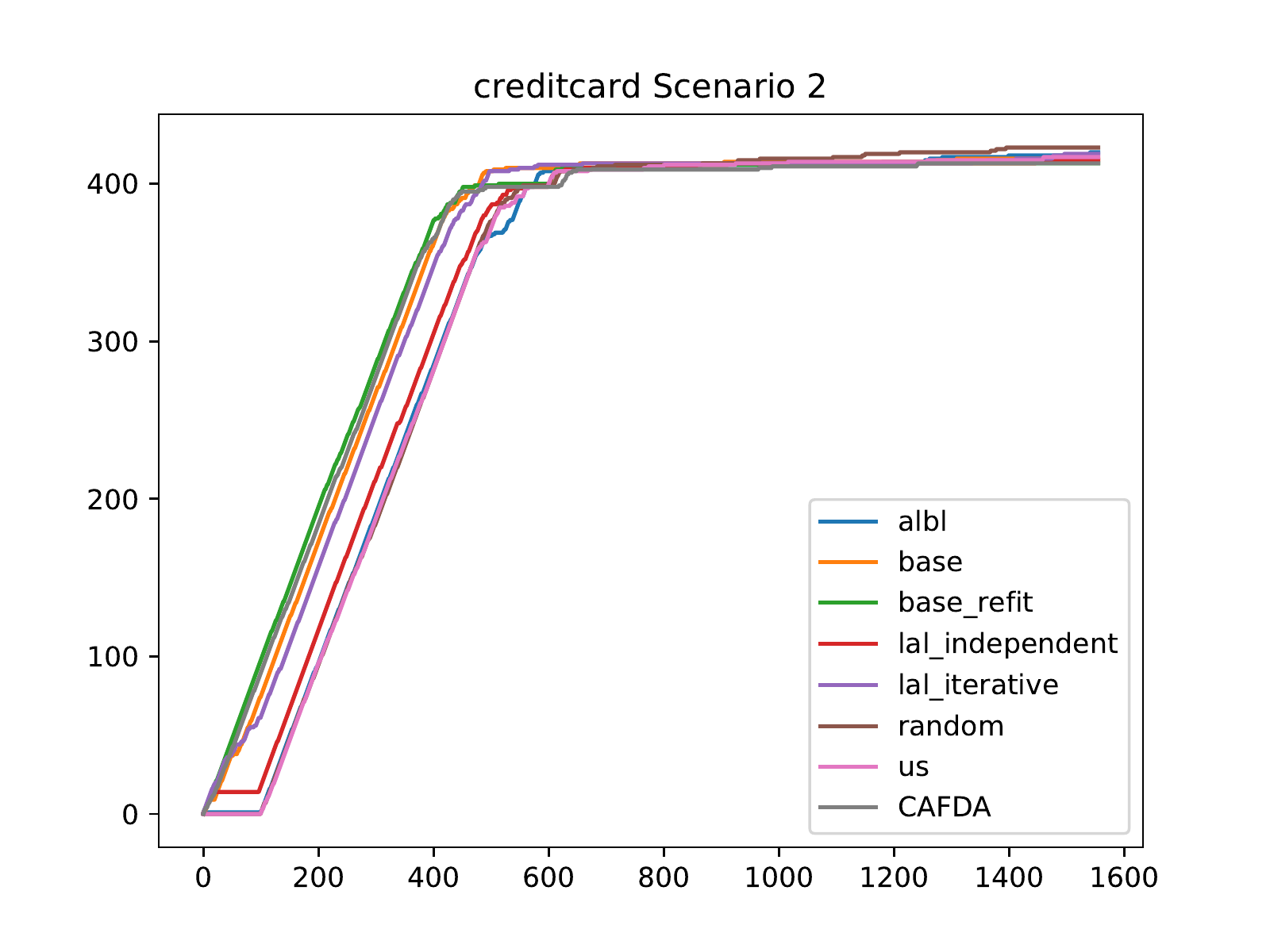} }}
    \subfloat[credit card scenario 2 first iterations ]{{\includegraphics[width=8cm]{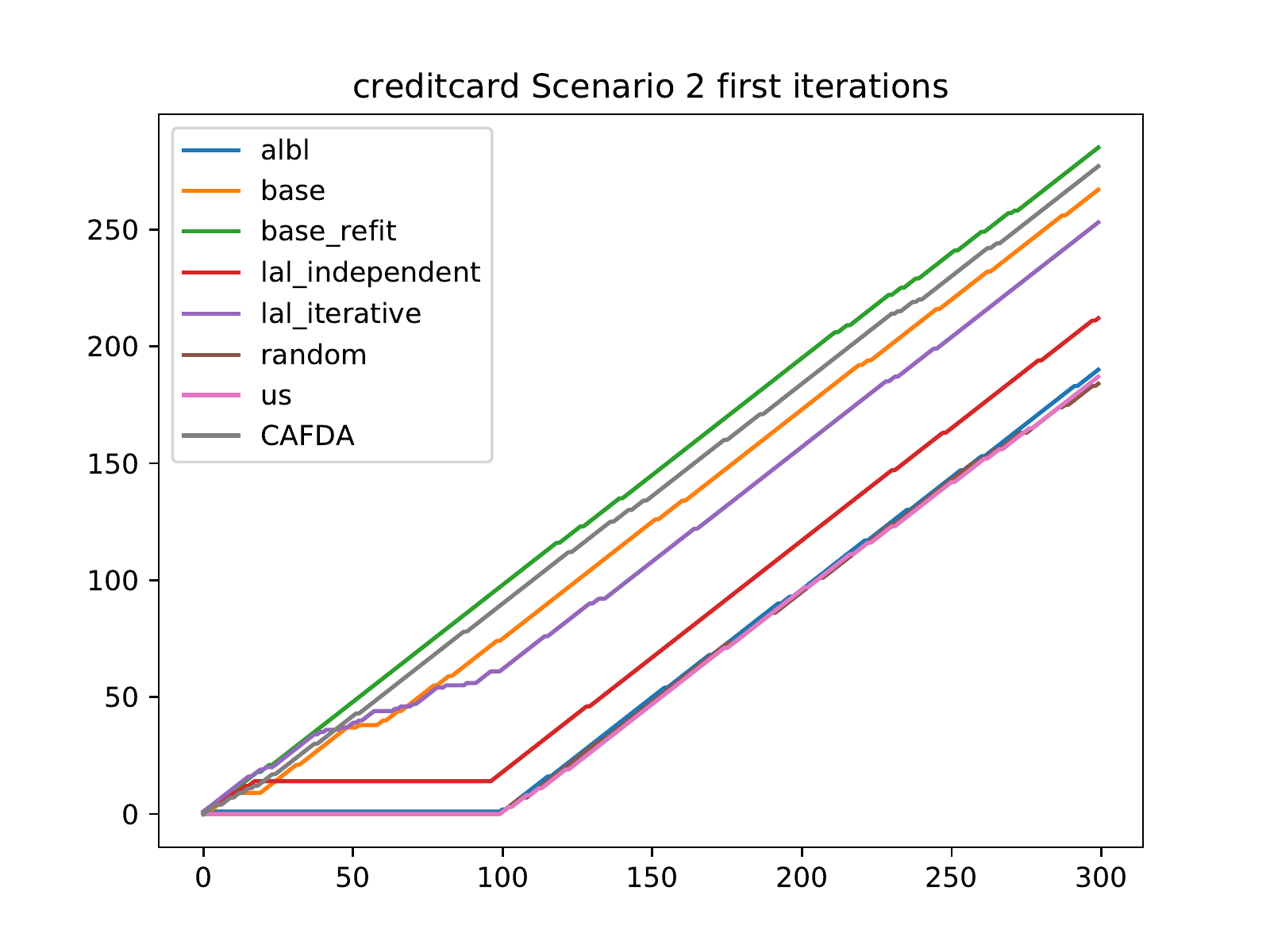} }}
    \caption{Scenario 2 cumulated rewards total and first iterations}
    \label{fig_2:scenario2}%
\end{figure}

We now turn to the case where each active learning strategies is used for the 100 first steps. In the next iterations, the points provided to the oracle are queried by solving $x^\star = \argmax_x \hat{p}(y = 1 | x)$ based on the resulting learned predictor. The results presented in Figure \ref{fig_2:scenario2} show that the active learning strategies do not take advantage even lately of their early exploration. Indeed \texttt{CAFDA}, \texttt{base} and \texttt{base\_refit} remain competitive while being simpler than active learning procedures. We focused on the 300 first iterations
where we observe that the delay of the cumulated reward of active learning procedures is generated at the very beginning and remains present until all the 1 have been found. 

\section{Conclusion and perspectives}
We presented a new fraud detection framework that differs from the active learning setting in which the quality of a strategy is measured by its ability to retrieve the rare labels to the oracle. We have shown that our algorithm \texttt{CAFDA} and also simple baselines provide better results than state of the art active learning algorithms on these problems. Future work will focus on the statistical properties of the computer-assisted fraud detection problem in order to design theory grounded optimal algorithms for the task at hand and will explore how to adapt other adaptative active learning strategies to our setting.



\bibliographystyle{apalike}
\bibliography{sample}

\end{document}